\documentclass{article}

\usepackage[sglblindworkshop, final]{paper}

\usepackage[utf8]{inputenc} % allow utf-8 input
\usepackage[T1]{fontenc}    % use 8-bit T1 fonts
\usepackage{hyperref}       % hyperlinks
\usepackage{url}            % simple URL typesetting
\usepackage{booktabs}       % professional-quality tables
\usepackage{amsfonts}% blackboard math symbols
\usepackage{subcaption} % for subfigure environment
\usepackage{enumitem}       % customized lists
\usepackage{nicefrac}       % compact symbols for 1/2, etc.
\usepackage{microtype}      % microtypography
\usepackage{xcolor}         % colors
\usepackage{amsmath}
\usepackage{graphicx}
\usepackage{natbib}

\title{Learning Without Critics? Revisiting GRPO in Classical Reinforcement Learning Environments}

\author{%
  \textbf{Bryan L. M. de Oliveira}$^{1,3,}$\thanks{Equal contribution. Correspondence to Bryan L. M. de Oliveira <bryanlincoln@discente.ufg.br>.} \quad
  \textbf{Felipe V. Frujeri}$^{1,2*}$ \quad
  \textbf{Marcos P. C. M. Queiroz}$^{1,3,*}$ \\
  \textbf{Luana G. B. Martins}$^{1}$ \quad
  \textbf{Telma W. de L. Soares}$^{1,3}$ \quad
  \textbf{Luckeciano C. Melo}$^{1,4}$ \quad
  \\\\
  $^1$Advanced Knowledge Center for Immersive Technologies – AKCIT \quad
  $^2$NVIDIA \\
  $^3$Institute of Informatics, Federal University of Goiás \quad
  $^4$OATML, University of Oxford
}

\begin{document}

\maketitle

% TLDR: We evaluate GRPO in classical RL, finding learned critics essential for long-horizon tasks, high discount factors generally optimal when early termination aids learning, and smaller groups outperforming larger ones.

% Around 6 sentences:
% setting, why should we care?, problem, why solving is relevant?, our solution, results/implications
\begin{abstract}
Group Relative Policy Optimization (GRPO) has emerged as a scalable alternative to Proximal Policy Optimization (PPO) by eliminating the learned critic and instead estimating advantages through group-relative comparisons of trajectories. This simplification raises fundamental questions about the necessity of learned baselines in policy-gradient methods. We present the first systematic study of GRPO in classical single-task reinforcement learning environments, spanning discrete and continuous control tasks. Through controlled ablations isolating baselines, discounting, and group sampling, we reveal three key findings: (1) learned critics remain essential for long-horizon tasks—all critic-free baselines underperform PPO except in short-horizon environments like CartPole where episodic returns can be effective; (2) GRPO benefits from high discount factors ($\gamma = 0.99$) except in HalfCheetah, where lack of early termination favors moderate discounting ($\gamma = 0.9$); (3) smaller group sizes outperform larger ones, suggesting limitations in batch-based grouping strategies that mix unrelated episodes. These results reveal both the limitations of critic-free methods in classical control and the specific conditions where they remain viable alternatives to learned value functions.
\end{abstract}

\section{Introduction}

% Around 4 paragraphs:

% setting, why should we care? why solving is relevant?

Policy gradient methods form the backbone of modern reinforcement learning, directly optimizing policies to maximize expected rewards \cite{williams1992simple,sutton1999policygradient}. Among these, Proximal Policy Optimization (PPO) \cite{schulman2017ppo} has emerged as a dominant approach, combining stability with sample efficiency through its use of a learned value function (critic) and clipped objective. However, the computational overhead of maintaining both actor and critic networks has motivated the development of critic-free alternatives, particularly in resource-constrained settings such as large language model (LLM) training.

% problem, main related work and literature gap

Group Relative Policy Optimization (GRPO) represents one such alternative, originally developed for reinforcement learning from human feedback (RLHF) in LLMs \cite{shao2024grpo}. GRPO eliminates the critic entirely, instead estimating advantages by comparing returns across groups of trajectories sampled from the same state or prompt. While this approach has shown promise in language model fine-tuning, its behavior in classical reinforcement learning environments remains largely unexplored.

The theoretical foundation for baseline methods in policy gradients is well-established. REINFORCE \cite{williams1992simple} introduced the use of baselines to reduce variance without introducing bias, where the gradient estimator subtracts a baseline $b(s)$ from the return $R_t$. Common baseline choices include scalar averages, learned state-value functions, or advantage functions. The latter approach led to the development of actor-critic methods like A2C and A3C \cite{mnih2016a3c}. PPO builds upon this foundation by using a learned critic to estimate advantages, typically combined with Generalized Advantage Estimation (GAE) for bias-variance control. Its clipped objective prevents destructively large policy updates while maintaining sample efficiency. GRPO replaces the learned critic with a group-based baseline computed from episode returns. Given a group of $G$ sampled episodes from the same initial state, GRPO normalizes the returns across the group to estimate advantages. This group-relative normalization removes the need for a learned value function while providing a baseline that adapts to the current policy's performance distribution.

% our solution

This work addresses a fundamental question in policy gradient methods: \emph{How does GRPO compare against PPO and its variants in standard RL benchmarks?} We conduct the first systematic evaluation of GRPO across classical reinforcement learning environments, spanning discrete control (CartPole, Acrobot), and continuous control (MountainCarContinuous, HalfCheetah, Humanoid). Our study is designed around controlled ablations that examine three key differences: (1) baseline estimation, comparing GRPO's group-relative normalization against PPO's learned critic and simpler alternatives; (2) horizon and discounting, analyzing GRPO's preference for $\gamma=1$ across different task structures; and (3) grouping strategies, evaluating how group sizes affect performance.

% results/implications/contributions

Our findings reveal that GRPO's effectiveness is highly context-dependent in classical RL settings. We show that critic-free baselines cannot replace learned value functions in long-horizon continuous control tasks—GRPO performs comparably to simple batch baselines but substantially underperforms PPO across all environments except CartPole. The presence of early termination emerges as a critical factor determining when critic-free methods can extract meaningful learning signals. GRPO generally benefits from high discount factors ($\gamma = 0.99$ often optimal), with notable exceptions in environments lacking early termination like HalfCheetah (which favors $\gamma = 0.9$). Smaller group sizes surprisingly outperform larger ones, even when controlling for update frequency, suggesting limitations in batch-based grouping strategies that mix potentially unrelated episodes from different parallel environments. These results establish the first systematic evaluation of GRPO beyond language modeling, revealing both its limitations and the conditions under which critic-free methods remain viable.

To facilitate reproducibility, we provide code, configuration files, and experiment logs that reproduce all results reported in this paper at \url{https://github.com/AKCIT-RL/revisiting-grpo}.

\section{Related Work}

% Policy Gradent Methods, exploration of baselines, PPO and GRPO
% GRPO and Classical RL Settings
% Empirical and Theoretical Study of Baselines

\textbf{Policy Gradient Methods and Baselines.}
Policy gradient methods originated with REINFORCE \cite{williams1992simple}, with the policy gradient theorem formalized for function approximation \cite{sutton1999policygradient}. Because Monte Carlo returns have high variance, baselines act as control variates to reduce variance without bias \cite{williams1992simple,greensmith2004variance}. The optimal state-dependent baseline minimizing variance was shown to be the value function itself \cite{weaver2001optimal}. Recent work also revisits the original REINFORCE algorithm proposing alternative gradient estimation techniques \cite{bhatnagar2023reinforce}. Actor-critic methods (e.g., A2C/A3C \cite{mnih2016a3c}, NAC \cite{peters2008nac}) learn $V(s)$ to produce state-dependent advantages; GAE further improves bias--variance trade-offs \cite{schulman2018gae}. Beyond value baselines, control-variates such as Q-Prop and action-dependent baselines also target variance in continuous control \cite{gu2017qprop,wu2018adb}, though more complex action-dependent baselines do not necessarily reduce variance over standard state-dependent baselines in practice \cite{tucker2018mirage}. To stabilize updates, trust-region and natural-gradient methods \cite{kakade2001npg} culminate in TRPO \cite{schulman2017trpo}, with PPO's clipped surrogate providing a first-order approximation to conservative updates \cite{kakade2002cpi,schulman2017ppo}.

\textbf{RL in Large Language Models.}
With the advancement of Large Language Models (LLMs), the challenge of aligning their results with human intentions arises. RLHF fine-tunes pretrained LMs using KL-regularized PPO-like objectives \cite{christiano2017preferences,ziegler2019finetuning,stiennon2020summarize,ouyang2022instructgpt}. PPO is typically adapted to start from a supervised fine-tuned (SFT) reference policy with a KL control term to regularize updates, stabilizing optimization in high-dimensional action spaces.

Due to its robustness, PPO quickly became the default algorithm for the optimization step in the RLHF pipeline \cite{ziegler2020llmrlhf}. In the case of language models, this burden is amplified, as PPO training typically requires four models in memory: the actor (policy), the critic (value function), a frozen reference policy for KL regularization, and the reward model, although in practice the actor and critic may share parameters to reduce overhead. It was in this context that Group Relative Policy Optimization (GRPO) \cite{shao2024grpo} was introduced, proposing a simpler approach.

Several alternative approaches address the computational and implementation challenges of PPO in LLMs. Direct Preference Optimization (DPO) \cite{rafailov2023dpo} and Sequence Likelihood Calibration (SLiC-HF) \cite{zhao2023slichf} reformulate alignment as contrastive learning problems, training the model to assign higher likelihood to preferred sequences over dispreferred ones without the need for rollouts against a critic or a frozen reference policy. Reinforcement Learning from AI Feedback (RLAIF) \cite{bai2022constitutional} extends the alignment paradigm by replacing human-labeled preference data with synthetic feedback generated by large language models themselves, further reducing the cost and memory requirements of training. Collectively, these methods share the goal of simplifying the optimization pipeline, reducing reliance on multiple large neural networks, and lowering the memory footprint inherent in PPO-based RLHF.

\textbf{GRPO and Applications.}
Group Relative Policy Optimization (GRPO) \cite{shao2024grpo} breaks with the historical trend of adding complexity to stabilize training by completely removing the need for a learned critic—a known source of instability and bias due to function approximation errors, and often requiring careful tuning of numerous implementation details for peak performance \cite{andrychowicz2020whatmatters}. Instead, it introduces a simpler baseline: for each prompt, multiple responses are sampled, and the average return of that group is used to normalize the advantages of the individual responses. This design not only reduces computational and memory overhead but also makes GRPO particularly attractive in the context of reasoning-focused language models and long-context settings, where generating and storing many rollouts already imposes significant resource demands.

Although born in the language domain, GRPO has demonstrated potential for generalization beyond language modeling. Recent work has explored GRPO in multi-task robotics \cite{joshi2025mtbench}, where trajectories are grouped by task, and in continuous control \cite{khanda2025extending}, where k-means clustering is used to group trajectories, adding significant computational overhead. However, these studies do not systematically investigate the impact of key implementation choices—such as baseline selection, discount factors, and group size—on GRPO's performance. GRPO moves in the opposite direction of the common trend in RL by simplifying the learning algorithm, eliminating the learned critic—a known source of instability and bias due to function approximation errors. This simplification leads to the central questions of our work: Is this approach more robust than its more complex counterparts? Is GRPO a general-purpose algorithm, or is its effectiveness specific to certain domains? We therefore investigate whether this simpler approach can be a viable alternative for a broader class of RL problems, with particular attention to the algorithmic components that determine its success.

\section{Background}\label{sec:background}

We consider a discounted Markov decision process (MDP) with state space $\mathcal{S}$, action space $\mathcal{A}$, transition kernel $P$, reward function $r$, and discount $\gamma \in [0,1]$. A stochastic policy $\pi_\theta(a\mid s)$ maximizes the expected discounted return $J(\theta) = \mathbb{E}_{\pi_\theta} \big[ \sum_{t=0}^{\infty} \gamma^t r_t \big]$.

The score-function estimator yields the policy gradient \cite{williams1992simple,sutton1999policygradient}
\begin{equation}
    \nabla_\theta J(\theta) = \mathbb{E}_{\pi_\theta} \Bigg[ \sum_{t=0}^{\infty} \nabla_\theta \log \pi_\theta(a_t\mid s_t)\, \big( R_t - b(s_t) \big) \Bigg],
\end{equation}
where $R_t = \sum_{l\ge 0} \gamma^l r_{t+l}$ is the Monte Carlo return and $b(s_t)$ is any baseline independent of $a_t$. Subtracting $b(s_t)$ preserves unbiasedness while reducing variance via a control variate \cite{greensmith2004variance}.

In actor-critic methods, the baseline is typically a learned value function $b(s_t)=V_\phi(s_t)$, making the term $(R_t - V_\phi(s_t))$ an estimate of the advantage function $A(s_t,a_t)=Q(s_t,a_t)-V(s_t)$. To reduce variance further, bootstrapping replaces the high-variance Monte Carlo return $R_t$ with the lower-variance estimate $r_t + \gamma V_\phi(s_{t+1})$, yielding the temporal difference error $\delta_t = r_t + \gamma V_\phi(s_{t+1}) - V_\phi(s_t)$, which approximates the advantage $A(s_t,a_t)$. Generalized Advantage Estimation (GAE) \cite{schulman2018gae} extends this by combining multiple temporal difference errors: $\hat{A}_t = \sum_{l=0}^{L-1} (\gamma\lambda)^l \, \delta_{t+l}$, where $L$ is the rollout horizon and $\lambda\in[0,1]$ controls the bias--variance trade-off.

Proximal Policy Optimization (PPO) \cite{schulman2017ppo} stabilizes on-policy updates by clipping the importance ratio $r_t(\theta)=\frac{\pi_\theta(a_t\mid s_t)}{\pi_{\theta_{\text{old}}}(a_t\mid s_t)}$, a first-order approximation to conservative policy iteration \cite{kakade2002cpi}. Given an advantage estimate $\hat A_t$ (typically from GAE), PPO maximizes the clipped surrogate
\begin{equation}
    \mathcal{L}_{\text{clip}}(\theta) = \mathbb{E}\, \big[ \min\big( r_t(\theta)\, \hat A_t,\; \operatorname{clip}(r_t(\theta),\,1-\epsilon,\,1+\epsilon)\, \hat A_t \big) \big],
\end{equation}
typically combined with a value loss and (optionally) entropy regularization.

Group Relative Policy Optimization (GRPO) \cite{shao2024grpo} takes a fundamentally different approach by eliminating the learned critic entirely. Instead of using bootstrapped temporal difference errors, GRPO returns to Monte Carlo returns $R(\tau_i)$ for entire episodes $\tau_i$ and computes a group-relative baseline. For a group of $G$ trajectories sampled from a common initial context, GRPO normalizes each episode's return relative to the group statistics:
\begin{equation}
    \hat{A}^{\text{GRPO}}(\tau_i) = \frac{R(\tau_i) - \mu_G}{\sigma_G + \epsilon},\quad
    \mu_G = \frac{1}{G}\sum_{j=1}^G R(\tau_j),\quad
    \sigma_G^2 = \frac{1}{G}\sum_{j=1}^G (R(\tau_j) - \mu_G)^2.
\end{equation}
This scalar advantage is then applied to all time steps within episode $\tau_i$. In language settings, the common initial context is a prompt; in this paper, we form groups for classical RL tasks from trajectories collected in parallel within the same batch.

The key simplification in GRPO is the replacement of PPO's step-wise, bootstrapped advantages with episode-wise, group-relative advantages. This eliminates the need for a learned value function but sacrifices the temporal granularity and bias-variance control that GAE provides. The group-relative normalization serves as a variance-reducing baseline, but unlike PPO's learned critic, it cannot adapt to state-specific value estimates or provide credit assignment within episodes.

We distinguish the environment horizon $H$ from the rollout horizon $N_{\text{steps}}$. Unless stated otherwise, we use $N_{\text{steps}}=128$ for standard PPO and $N_{\text{steps}}=H$ for variants without a value function. The discount factor $\gamma$ modulates credit assignment across time; GRPO is often used with $\gamma=1$, while PPO commonly uses $\gamma<1$ with a learned critic.

\section{Central Research Question and Motivation}

\begin{figure}[t]
    \centering
    \includegraphics[width=\linewidth]{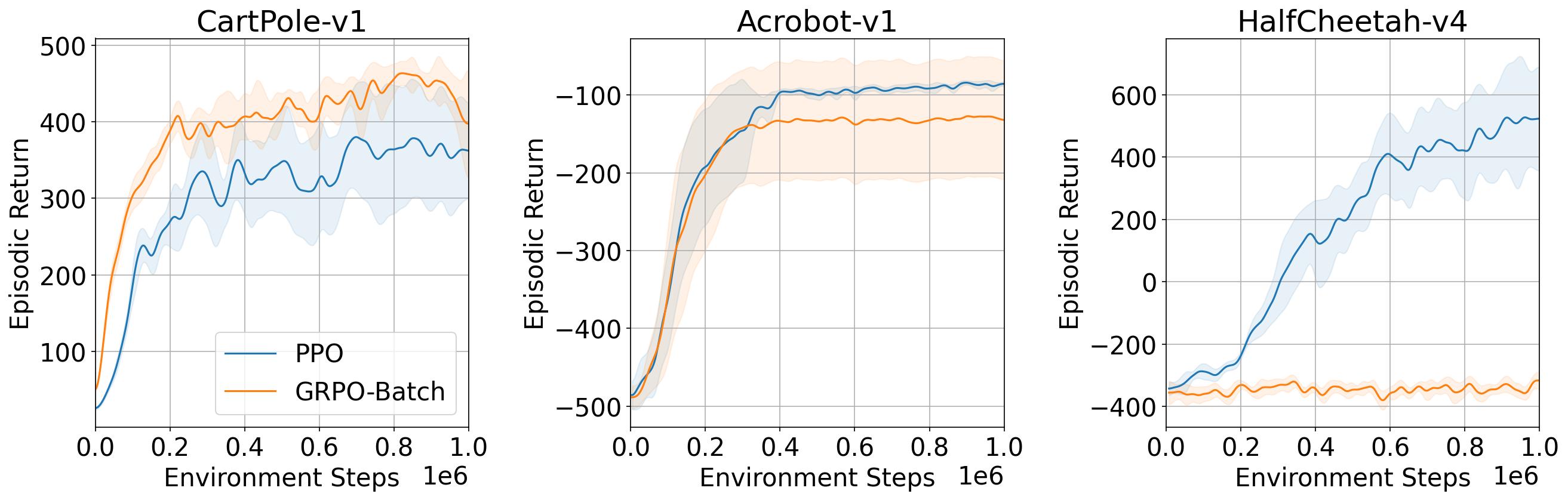}
    \caption{\textbf{Performance comparison of PPO and GRPO with commonly used settings.} This motivating example shows PPO ($\gamma=0.99$, $N_{\text{steps}}=128$) and GRPO ($\gamma=1$, $N_{\text{steps}}=H$) using our grouping strategy. Relative performance varies and depends on the environment characteristics.}
    \label{fig:ppo_vs_grpo}
\end{figure}

As illustrated in Figure~\ref{fig:ppo_vs_grpo}, the relative performance between PPO and GRPO varies significantly across different environments, making it unclear which algorithmic components or environment characteristics define GRPO's effectiveness and how hyperparameters influence learning dynamics. We study whether a critic-free policy-gradient method can match PPO in classical single-task RL when its variance-reduction mechanism is purely group-relative. Our central question is: \emph{under which conditions does GRPO provide stable and sample-efficient learning compared with PPO's clipped policy updates and learned critic?}

We focus on three factors that govern the quality of policy gradients in this setting:
\begin{enumerate}[leftmargin=*]
    \item \textbf{Baselines and advantage estimation:} How do different baselines (none, batch statistics, and learned value functions) affect stability and data efficiency when combined with clipping?
    \item \textbf{Discounting and horizon:} How does the choice of $\gamma$ interact with GRPO's episodic, critic-free advantages, especially in long-horizon or dense-reward tasks where credit assignment is difficult?
    \item \textbf{Group sampling:} How does group size influence variance reduction and update frequency when groups are formed from parallel rollouts in classical RL?
\end{enumerate}

This study matters for two reasons. First, removing the critic simplifies the training stack and reduces memory and implementation complexity, which is attractive for large-scale or resource-constrained settings. Second, the absence of a learned baseline raises concrete risks: higher gradient variance, weaker credit assignment depending on $\gamma$ values, and ambiguity about how to define groups outside prompt-based domains. We therefore design controlled ablations that vary the baseline, the discount factor, and the group size while holding architectures, optimizers, rollout budgets, and clipping constants fixed. Our goal is to produce actionable guidance on when critic-free training suffices and when a learned value function remains necessary.

\section{Experimental Setup}

We evaluate PPO and GRPO under a unified protocol that isolates three factors derived from the policy-gradient formulation in Section~\ref{sec:background}: baselines, discounting and horizon, and group sampling.

\textbf{Environments and rollouts.} We use CartPole-v1, Acrobot-v1, MountainCarContinuous-v0, HalfCheetah-v4, and Humanoid-v4 \cite{towers2024gymnasium}, described in Appendix~\ref{ap:environments}. For each update, we collect rollouts with $N_{\text{envs}}=8$ parallel environments.

\textbf{Policies and optimization.} We keep network architectures, optimizers, learning rates ($2.5\times10^{-4}$), clipping coefficient ($\epsilon=0.2$), epochs per iteration ($N_{\text{epochs}}=4$), and minibatches per epoch ($N_{\text{mb}}=1$, to reduce stochasticity from sampling) fixed across ablations, sourcing default hyperparameters from CleanRL \citep{huang2022cleanrl}. Standard PPO uses a value head and GAE with $\lambda=0.95$, $\gamma=0.99$ and rollout length $N_{\text{steps}}=128$. GRPO uses the group-relative episodic advantage we define in Section~\ref{sec:rq1}, with groups of size 8, $\gamma=1$ and rollout length $N_{\text{steps}}=H$.

\textbf{Grouping in classical RL.} In the absence of prompts, we form groups from the $G$ trajectories collected concurrently from parallel environments within a batch. We apply the scalar group-relative advantage to all steps of each trajectory. This batch-based grouping strategy is computationally efficient compared to maintaining a learned value function, as in the original GRPO \citep{shao2024grpo}. We leave the exploration of more complex grouping strategies (e.g., similarity-based or task-based clustering) for future work, noting that such approaches may introduce computational overhead that could exceed the cost of maintaining a value function.

\textbf{Evaluation protocol.} We run 10 seeds per configuration and report the mean episodic return with 95\% confidence interval. All agents are trained with a budget of 1M environment steps. For readability, we display smoothed learning curves using a 1D Gaussian filter ($\sigma=5$).

% \item \textbf{Exploration Incentive}: the difference in exploration between PPO, which uses a policy entropy bonus, and GRPO, which has an implicit incentive through multiple  samples per state. Our goal is to test whether the rapid entropy collapse in GRPO can hinder exploration when the reference policy is not random.

% \item \textbf{KL Penalty}: PPO typically uses a KL penalty to restrict policy change, while GRPO uses DKL for a reference policy. We will investigate how this difference  affects training and whether using a random reference policy in GRPO can be reduced to an entropy bonus.

% \item \textbf{Compute Requirements / Hyperparameter Tuning}: Finally, we will compare the compute requirements and hyperparameter stability of both algorithms. We hypothesize  that, although GRPO requires less memory due to its lack of a criticality, it may require a larger group size to achieve the same sampling efficiency as PPO, which could result  in a higher total number of operations for convergence.

\section{RQ1: Do critic-free baselines match PPO's learned critic in classical RL?}
\label{sec:rq1}

A core difference between PPO and GRPO lies in their choice of baseline for variance reduction. PPO employs a learned value function $V_\phi(s)$, typically combined with GAE, to compute a low-variance, bias-controlled estimate of the advantage. GRPO, in contrast, removes the critic entirely and uses the group-relative episodic advantage defined in Section~\ref{sec:background}. This group-based baseline connects directly to the design of reward modeling pipelines in LLMs, where sets of responses are compared within a shared prompt context. However, in classical RL tasks, the role of group baselines in mitigating high variance is still underexplored.
% Beyond learned critics, control-variate baselines further reduce gradient variance in continuous control \cite{greensmith2004variance,gu2017qprop,wu2018adb}. 

We ask whether critic-free baselines can match PPO's learned value function in stability and sample efficiency. We compare variants that differ only in how they compute advantages while keeping clipping and optimization fixed.

% \textbf{Research Questions.}  
% We pose the following questions:
% \begin{itemize}
%     \item How does the absence of a learned critic in GRPO affect the variance and bias of gradient estimates in long-horizon control tasks?
%     \item Can group size $G$ and baseline normalization reduce instability, and under what conditions?
%     \item How does clipping interact with different baseline definitions, especially when variance is high?
% \end{itemize}

% \textbf{Hypotheses.}  
% We hypothesize that:
% \begin{enumerate}[label=H\arabic*]
%     \item GRPO exhibits higher gradient variance than PPO in tasks with long horizons or delayed rewards, leading to unstable learning.  
%     \item Increasing $G$ or normalizing the group baseline reduces this instability.  
%     \item The group baseline is most effective when initialized from a strong prior policy, analogous to pretrained LLMs.  

\textbf{Hypotheses.} In long-horizon tasks, critics reduce variance more effectively than critic-free estimators. Group-relative normalization approximates batch baselines: it reduces scale sensitivity but cannot replace state-dependent credit assignment.
% Increasing $G$ or normalizing the group baseline reduces this instability.  
% Benefits of group-relative baselines grow when starting from a strong prior, as in language settings.

\textbf{Experimental design.} We compare PPO-style clipping paired with the following baselines:
\begin{enumerate}%[label=E\arabic*]
    \item \textbf{Standard PPO:} value function + GAE ($\lambda=0.95$), $\gamma=0.99$ and $N_{\text{steps}}=128$.
    \item \textbf{PPO w/o Baseline:} reduces to REINFORCE with multiple epochs and clipping ratio.  
    \item \textbf{PPO + Random Gaussian:} subtracting a random scalar baseline sampled from $\mathcal{N}(\mu, \sigma^2)$ matching return distribution statistics.
    \item \textbf{PPO + EMA:} exponential moving average of episodic returns with a 0.9 multiplier.
    \item \textbf{PPO + Batch Mean:} subtracting the mean of the return distribution.
    \item \textbf{PPO + Batch Mean and Scaling (GRPO-Batch):} batch mean baseline with advantages scaled by standard deviation of the return distribution.  
    % \item \textbf{TD Baseline:} PPO with $\lambda=0$ (pure TD(0) baseline).  
    % \item \textbf{GRPO + Critic:} GRPO augmented with value function baseline, equivalent to PPO with $\lambda=0$.  
\end{enumerate}

\begin{figure}[t]
    \centering
    \includegraphics[width=\linewidth]{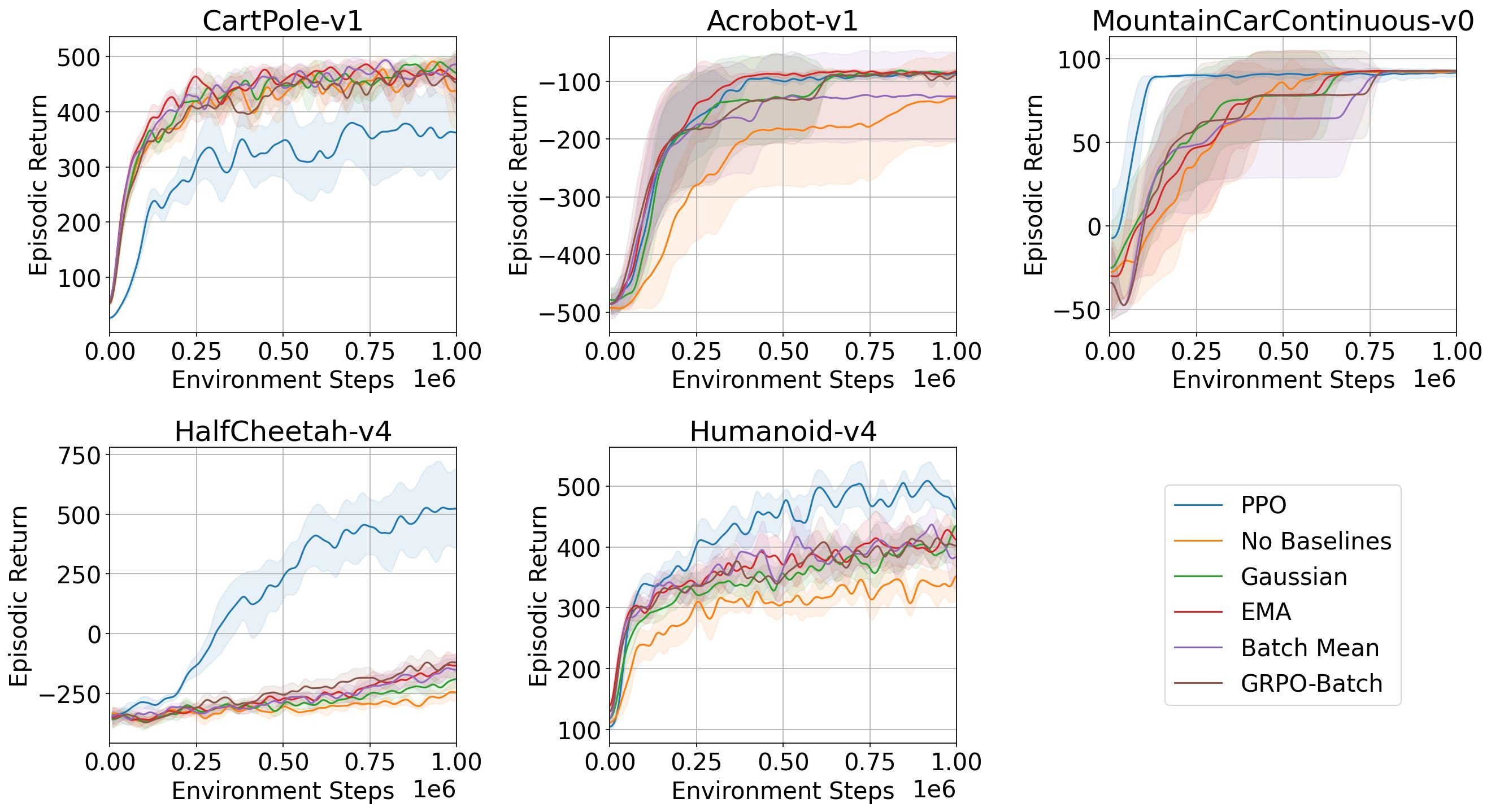}
    \caption{\textbf{Baseline ablations across environments.} We compare PPO with its learned critic against PPO variants without a baseline, with simple alternatives (batch mean, Gaussian, EMA), and GRPO with group-relative normalization. Removing the baseline substantially increases variance, especially in long-horizon continuous control, while GRPO performs comparably to simple baselines.}
    \label{fig:d1_grid}
\end{figure}

\textbf{Experiments and discussion.}
Figure~\ref{fig:d1_grid} shows the results with $\gamma=1$ and Monte Carlo returns for all variants (except standard PPO, which uses $\gamma=0.99$ and $N_{\text{steps}}=128$). All critic-free baselines exhibit similar performance, with most results falling within confidence bounds of each other. The key distinction is that PPO with its learned value function substantially outperforms all critic-free alternatives across all environments except CartPole.

In CartPole, however, the pattern reverses: critic-free methods often exceed standard PPO's performance. This occurs because CartPole's short horizon and simple dynamics make it particularly sensitive to overtraining. Our PPO implementation frequently reaches the maximum return of 500 but then experiences rapid performance degradation, likely due to overfitting or policy collapse from excessive updates. Since we do not perform extensive hyperparameter tuning (e.g., adjusting $N_{\text{epochs}}$ or $N_{\text{mb}}$), PPO's performance in this environment may be underrepresented. The critic-free methods, having higher variance, may benefit from implicit regularization that prevents such overtraining.

HalfCheetah presents a particularly revealing case: none of the critic-free baselines successfully learn with $\gamma=1$, requiring lower discount factors to achieve meaningful progress (see Figure~\ref{fig:d2_grid} in Section~\ref{sec:rq2}). This failure is likely attributable to HalfCheetah being the only environment in our suite without early termination conditions. With $\gamma=1$ and no episode boundaries, the learning signal becomes diluted across entire trajectories. In contrast, environments with early termination create clear separation between successful and unsuccessful trajectories, enabling critic-free baselines to extract meaningful learning signals even without state-dependent value estimates.

These observations highlight fundamental differences in how baselines address variance and credit assignment. A learned value function $V_\phi$ provides state-dependent baselines that reduce variance at each transition and enable precise credit assignment throughout an episode \cite{sutton1999policygradient,greensmith2004variance}. Episodic scalars from batch statistics or group-relative normalization can only centralize the return distribution—they cannot distinguish between transitions of different values within the same trajectory. This limitation becomes critical in environments without natural episode boundaries, where the absence of termination signals prevents the episodic advantage from effectively propagating credit. The connection to language model settings helps explain when critic-free methods succeed: RLHF scenarios typically feature sparse terminal rewards, clear episode boundaries, and strong pretrained policies. Classical control environments often lack both the natural episode structure and strong priors that make critic-free estimation viable.

\textbf{Answer to RQ1.} In long-horizon tasks without early termination, a learned critic remains necessary for stability and sample efficiency. All critic-free baselines perform similarly to each other but substantially below PPO with a learned value function, except in short-horizon tasks like CartPole where episodic advantages can be effective. The presence or absence of early termination emerges as a critical factor determining whether critic-free methods can extract meaningful learning signals.
 
\section{RQ2: How does the discount factor affect GRPO across horizons?}
\label{sec:rq2}

The discount factor $\gamma$ shapes how policy gradient methods balance immediate versus future rewards. While PPO typically uses $\gamma < 1$ (e.g., $0.99$), GRPO's original formulation implicitly assumes $\gamma = 1$, treating all timesteps equally. This reflects GRPO's origins in language model fine-tuning, where sparse terminal rewards benefit from uniform credit assignment.

We investigate how different discount factors affect GRPO's performance when advantages are episodic and critic-free, particularly across varying rollout horizons.

\textbf{Hypotheses.} Using $\gamma=1$ hinders credit assignment and reduces sample efficiency in long-horizon or dense/mixed reward environments when advantages are episodic and critic-free.

\textbf{Experimental design.} We sweep GRPO across $\gamma \in \{0, 0.1, 0.5, 0.95, 0.99, 1\}$ for all environments.

\begin{figure}[t]
    \centering
    \includegraphics[width=\linewidth]{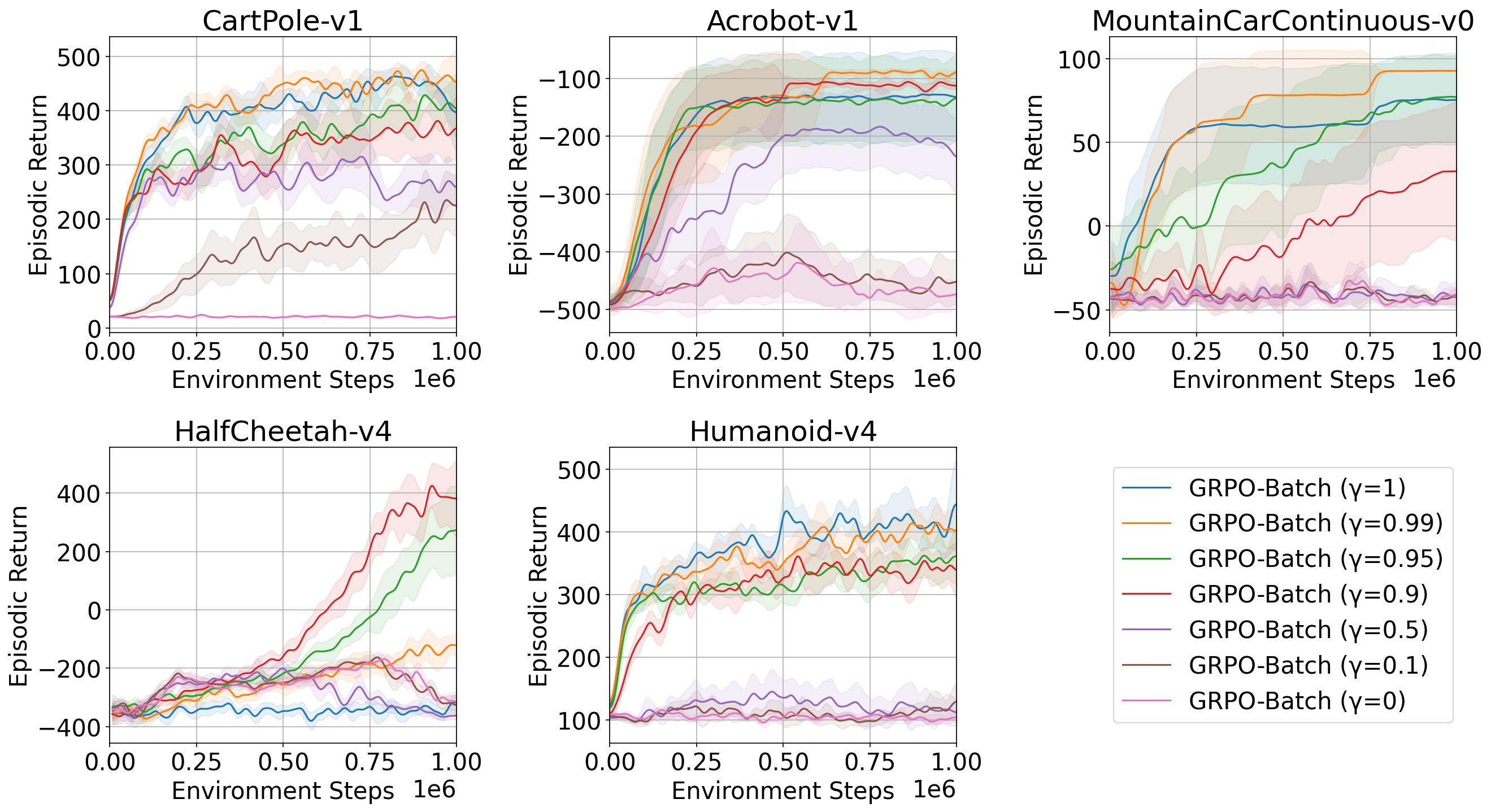}
    \caption{\textbf{Effect of discount factor across environments.} GRPO performance with varying $\gamma$ values. Higher $\gamma$ generally improves performance, with notable exceptions in HalfCheetah (optimal around $\gamma = 0.9$-$0.95$) and MountainCarContinuous (best at $\gamma = 0.99$).}
    \label{fig:d2_grid}
\end{figure}

\textbf{Experiments and discussion.}
Figure~\ref{fig:d2_grid} reveals that higher discount factors generally improve GRPO performance, with $\gamma = 0.99$ often yielding the best results. However, HalfCheetah presents a striking exception: $\gamma = 1$ produces the worst performance while $\gamma = 0.9$ achieves optimal results. This exception illuminates the interplay between discount factors and environment structure (task horizon and dynamics). HalfCheetah's unique characteristics—no early termination and temporally local dynamics—mean that with $\gamma = 1$, learning signals become diluted across entire trajectories, making it difficult to distinguish action quality. Moderate discounting ($\gamma = 0.9$) strikes the right balance, as very low discount factors are also undesirable since rewards are not immediate.

Contrary to our initial hypothesis, the impact of $\gamma = 1$ on credit assignment appears minimal in most environments. This likely stems from early termination creating natural separation between successful and unsuccessful trajectories, enabling episodic advantages to extract meaningful learning signals even without temporal discounting. This highlights a key difference between GRPO's natural domain (language modeling with sparse terminal rewards, where uniform credit assignment makes sense) and classical control: the presence or absence of early termination emerges as the critical factor determining whether high discount factors enable effective learning with episodic advantages.

\textbf{Answer to RQ2.} GRPO generally benefits from high discount factors, with $\gamma = 0.99$ often optimal across most environments. The notable exception is HalfCheetah, where its lack of early termination and temporally local dynamics favor moderate discounting ($\gamma = 0.9$). The minimal impact of $\gamma = 1$ on credit assignment in other environments likely stems from early termination providing natural episode structure that enables effective learning with episodic advantages.

\section{RQ3: How does group size influence stability and efficiency in GRPO?}
\label{sec:rq3}

A central distinction between PPO and GRPO lies in how trajectories are sampled and grouped. PPO typically samples one trajectory per environment per update, with each trajectory treated independently during advantage estimation. In contrast, GRPO relies on groups of episodes originating from the same initial state (or ``prompt'' in LLM terminology). These groups enable the computation of a relative baseline based on inter-episode comparisons. In classical RL environments, however, defining such groups is nontrivial, since resets produce stochastic initial states rather than fixed prompts. This raises questions about how to form effective groups and how group size affects stability and efficiency. In this work, we focus on batch-based grouping, where episodes are sampled from the same environment instance in parallel.

We investigate how group size affects stability and apparent sample efficiency when groups are formed from parallel environments within a batch.
% \item How sensitive are GRPO’s learning dynamics to different group sizes?
% \item What constitutes a suitable ``prompt-equivalent'' in classical RL tasks?

\textbf{Hypotheses.} Larger groups reduce the variance of the group baseline but decrease update frequency per environment step, affecting apparent sample efficiency. We expect larger groups to provide more stable learning through better baseline estimation, but at the cost of reduced update frequency.
% \item GRPO requires groups of trajectories with sufficiently similar prefixes to produce a meaningful baseline.
% \item Groups formed from identical initial states (same-seed resets) yield more stable training than groups formed from heterogeneous states.
% \item Larger group sizes provide more stable learning by contrasting good and bad rollouts, but increase sample complexity.

\textbf{Experimental design.} We assume $\gamma=0.99$ and vary the group size $G \in \{8, 16, 32, 64, 128\}$ while keeping other hyperparameters fixed.

\begin{figure}[t]
    \centering
    \includegraphics[width=\linewidth]{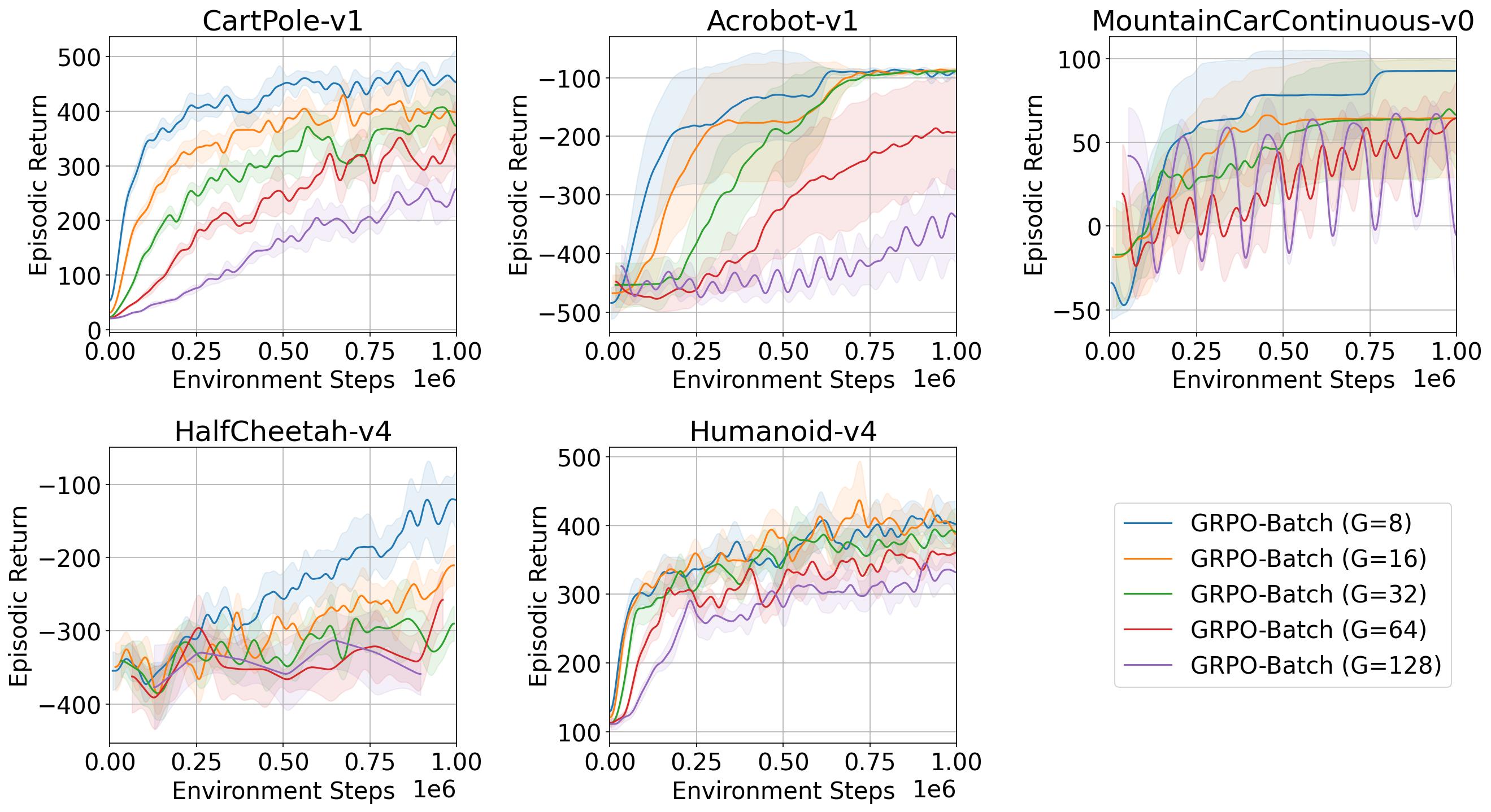}
    \caption{\textbf{Effect of group size on GRPO across environments.} Each subfigure shows episodic returns when varying the number of parallel environments used as a group ($G \in \{8, 16, 32, 64\}$).}
    \label{fig:d3_grid}
\end{figure}

% TODO add um número de trajetórias menor causa menos variancia no estimador
\textbf{Experiments and discussion.}
Figure~\ref{fig:d3_grid} reveals a surprising pattern: smaller groups of size 8 generally outperform larger ones across most environments. The superior performance of smaller groups initially appears counterintuitive, as larger groups should provide more stable baseline estimates by averaging over more trajectories. While increased update frequency per environment step partially explains this advantage, the trend largely persists even when we normalize performance by training iterations (see Appendix Figure~\ref{fig:group_size_h0_y0.99_iterations}).

We hypothesize that this advantage of smaller groups stems from our grouping approach, which bunches together episodes from different parallel environment instances. Unlike prompt-based grouping in language models where multiple responses share the same context, our strategy groups potentially unrelated episodes. In this setting, differences in episode returns may depend more on the specific states visited (which vary across parallel environments) than on policy quality differences that the group baseline should capture. Smaller groups may thus provide more coherent comparisons by reducing the heterogeneity of grouped episodes, even if at the cost of higher baseline variance.

These findings highlight an important limitation of our current grouping strategy and suggest that the design of effective grouping mechanisms for classical RL environments remains an open problem. Future work exploring alternative grouping strategies—such as grouping episodes with similar state distributions or trajectory characteristics—should investigate whether this dynamic persists or whether more sophisticated grouping can unlock the theoretical benefits of larger group sizes.

\textbf{Answer to RQ3.} Smaller groups tend to outperform larger ones in our experiments, with this advantage persisting even when controlling for update frequency. This likely reflects limitations in our grouping strategy, which mixes potentially unrelated episodes together. Alternative grouping approaches like \cite{khanda2025extending} that leverage trajectory similarity warrant investigation in future work.

\section{Conclusion}

In this paper, we present the first systematic study of GRPO in classical single-task RL environments. Through controlled ablations across discrete and continuous control tasks, we identify several key findings that explain GRPO's performance characteristics and limitations. First, we demonstrate that critic-free baselines cannot match PPO's learned value function in long-horizon tasks: while simple alternatives like batch mean and EMA provide some variance reduction, they lack the state-dependent credit assignment necessary for effective learning. The presence or absence of early termination emerges as a critical factor—environments with termination conditions create natural separation between successful and unsuccessful trajectories, enabling critic-free methods to extract meaningful learning signals, while those without (like HalfCheetah) fail to provide such differentiation. Second, our analysis reveals that GRPO generally benefits from high discount factors, with $\gamma = 0.99$ often optimal. The notable exception is HalfCheetah, where its lack of early termination and temporally local dynamics favor moderate discounting ($\gamma = 0.9$). Third, we find that smaller groups tend to outperform larger ones, even when controlling for update frequency, likely reflecting limitations in our batch-based grouping strategy which mixes potentially unrelated episodes together.

\textbf{Limitations and future work.} Our study has several limitations that suggest directions for future work. First, our batch-based grouping strategy may not effectively capture the benefits of larger group sizes seen in language model settings. Alternative strategies that leverage trajectory similarity or state distribution clustering warrant investigation, though computational overhead must be weighed against the cost of maintaining a value function. Second, future work should systematically investigate exploration mechanisms such as entropy regularization and other implementation details. The hyperparameter space involving epochs, mini-batches, clipping, learning rates, and normalization offers rich opportunities for optimization that we only partially explored. Finally, our evaluation focused on fully observable standard RL benchmarks; extending to partially observable environments or sparse reward domains may reveal different performance characteristics and trade-offs.

\begin{ack}
This work has been partially funded by the project Research and Development of Digital Agents Capable of Planning, Acting, Cooperating and Learning supported by Advanced Knowledge Center in Immersive Technologies (AKCIT), with financial resources from the PPI IoT/Manufatura 4.0 / PPI HardwareBR of the MCTI grant number 057/2023, signed with EMBRAPII
\end{ack}

\bibliographystyle{plain}
\bibliography{references}

%%%%%%%%%%%%%%%%%%%%%%%%%%%%%%%%%%%%%%%%%%%%%%%%%%%%%%%%%%%%
\newpage
\appendix

\section{Environments}\label{ap:environments}

To evaluate GRPO in comparison with REINFORCE and PPO, we conduct experiments across a diverse set of classical control and continuous control tasks. The chosen environments are selected to probe different aspects of variance reduction, exploration, and stability. All environments are implemented using the \texttt{Gymnasium} library \citep{towers2024gymnasium}.

\textbf{CartPole.}  
CartPole is a classic control task with a short horizon and binary termination signal. Its simplicity and fast runtime make it ideal for sanity checks and for quickly validating the correctness of our implementation. 
% We use this environment primarily to confirm that GRPO can match PPO performance in stable and well-understood settings before scaling up to more complex domains.

\textbf{Acrobot.}
Acrobot introduces a longer task horizon and sparse rewards, making it a suitable environment for investigating credit assignment. Specifically, we use it to test the hypothesis that setting $\gamma=1$ in GRPO may hinder credit assignment and degrade sample efficiency compared to PPO’s $\gamma < 1$.
% The environment thus serves as a benchmark for understanding how horizon length and discounting interact with group-relative normalization.

\textbf{MountainCarContinuous.}  
This task requires the agent to build momentum by swinging back and forth before reaching the goal. Because it demands sustained exploration, it is well-suited to test whether GRPO’s implicit exploration (via stochastic group sampling) is sufficient, or if an explicit entropy bonus is required for efficient learning.
% We use it to evaluate the hypothesis that the absence of entropy regularization in GRPO may hinder performance in exploration-heavy domains.

\textbf{HalfCheetah.}  
HalfCheetah is a standard MuJoCo locomotion task with continuous states and actions, and long horizons. This environment stresses both sample efficiency and stability, allowing us to investigate three main hypotheses: whether GRPO’s group-relative advantage leads to higher variance in long-horizon tasks and potentially destabilizes learning; whether GRPO requires larger group sizes to achieve comparable performance to PPO, thus increasing environment interactions and computational cost; and whether GRPO’s group-based sampling remains stable in high-dimensional control settings.

\textbf{Humanoid.}  
Finally, the standard Humanoid locomotion task represents one of the most complex benchmarks in continuous control, with high-dimensional states, continuous actions, and long horizons. We use it to evaluate the scalability of GRPO, testing whether group-relative normalization can remain competitive with PPO in high-dimensional settings, or if it degrades due to increased variance and sample inefficiency.  

\newpage
\section{Additional Results}

\subsection{Motivating example: PPO vs. GRPO on all environments}\label{ap:motivating_example}

\begin{figure}[h!]
    \centering
    \includegraphics[width=\linewidth]{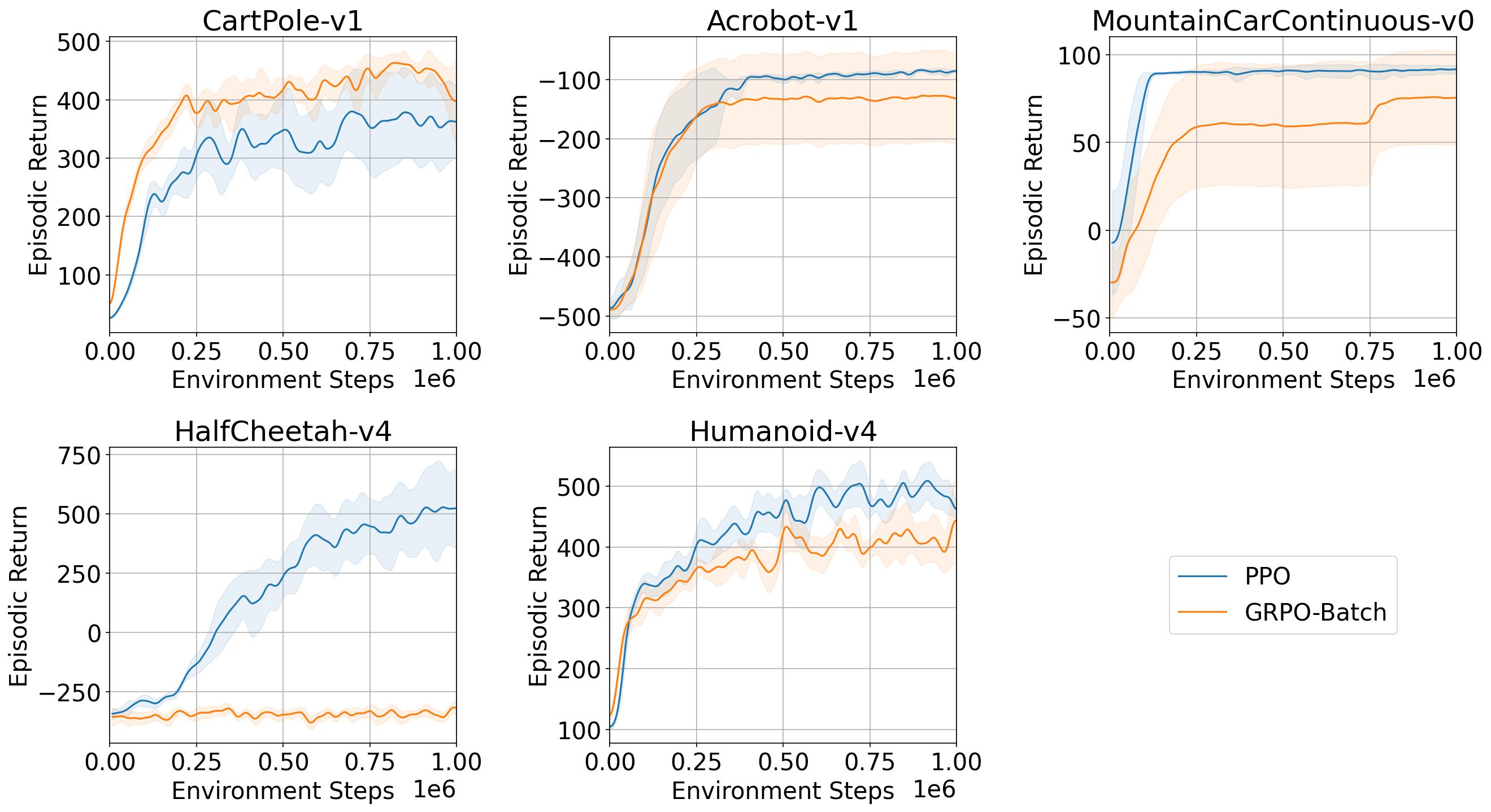}
    \caption{\textbf{PPO vs. GRPO.} This motivating example shows PPO with standard settings ($\gamma=0.99$, trajectory length 128) and GRPO with standard settings using our grouping strategy ($\gamma=1$, full episodes). Relative performance varies and depends on the environment characteristics. This figure shows all environments.}
    \label{fig:ppo_vs_grpo_full}
\end{figure}

\newpage
\subsection{PPO Across Discount Factors and Rollout Horizons}\label{ap:ppo_all_variations}

\begin{figure}[h!]
    \centering
    \includegraphics[width=\linewidth]{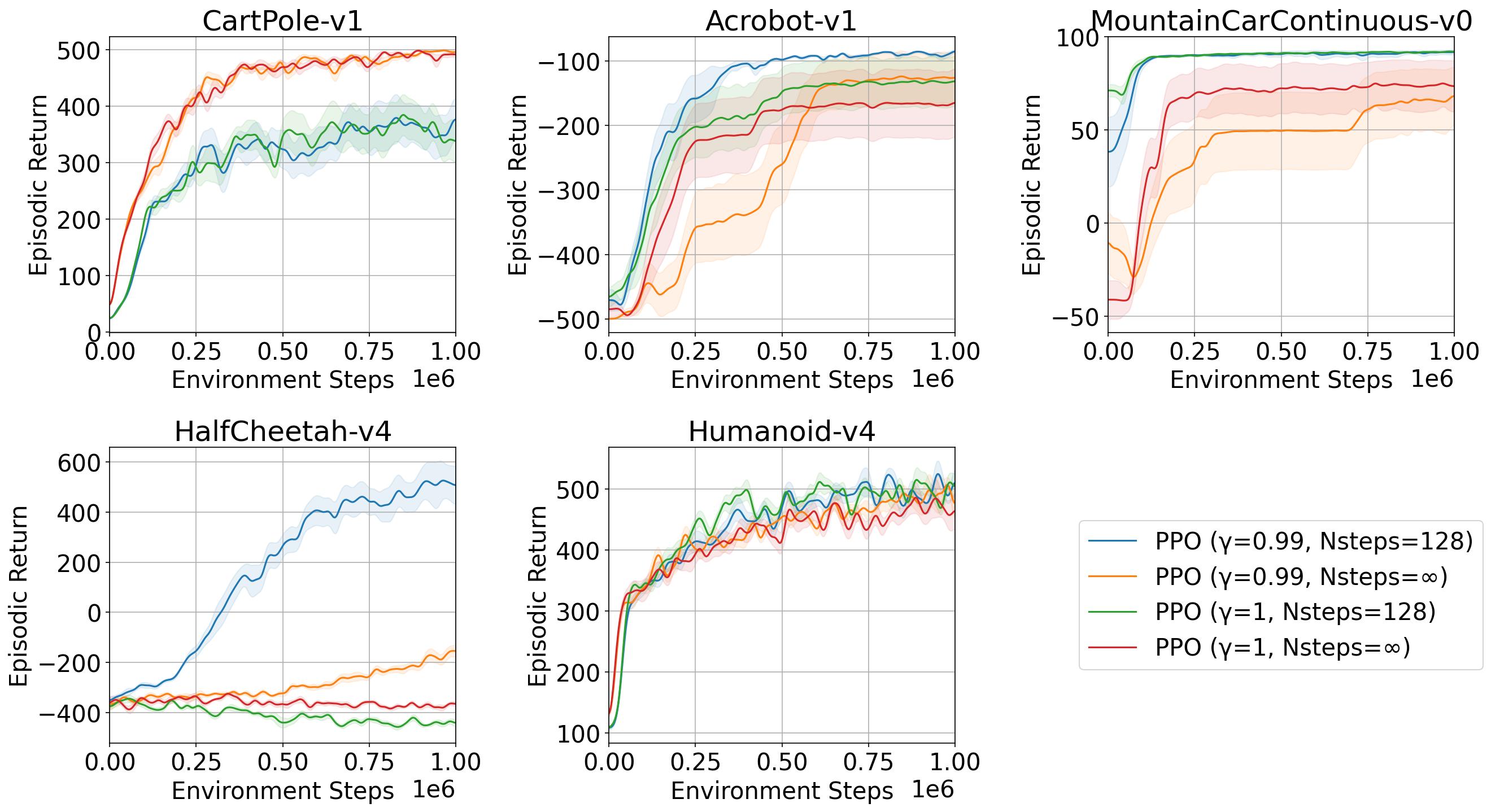}
    \caption{\textbf{PPO Variants.} This figure shows PPO variants with $\gamma \in \{0.99, 1\}$ and trajectory lengths in $\{128, \infty\}$ (using the value function for bootstrapping or not). Bootstrapping seems to hurt performance on CartPole, but is essential for HalfCheetah and beneficial for Acrobot. The choice of $\gamma$ has minor impact, with the lower value ($\gamma=0.99$) having better performance on HalfCheetah.}
    \label{fig:ppo_all_variations}
\end{figure}

\newpage
\subsection{Baselines Across Discount Factors}\label{ap:baselines_all_variations}

\begin{figure}[h!]
    \centering
    \includegraphics[width=\linewidth]{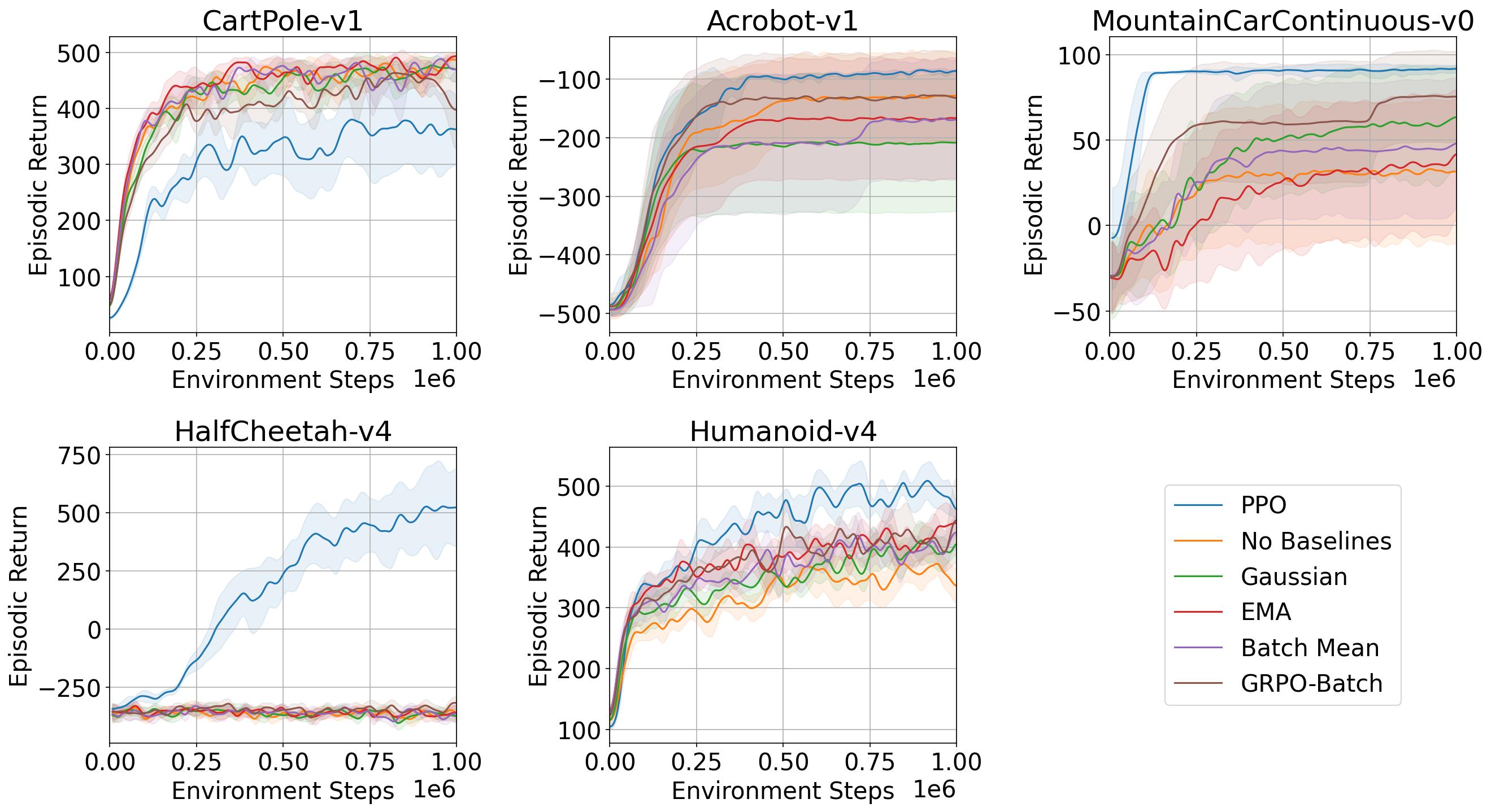}
    \caption{\textbf{Baseline ablations with $\gamma=1$.}}
    \label{fig:baselines_h0_y1}
\end{figure}

\newpage
\subsection{GRPO Across Group Size and Discount Factors}\label{ap:group_size}

\begin{figure}[h!]
    \centering
    \includegraphics[width=\linewidth]{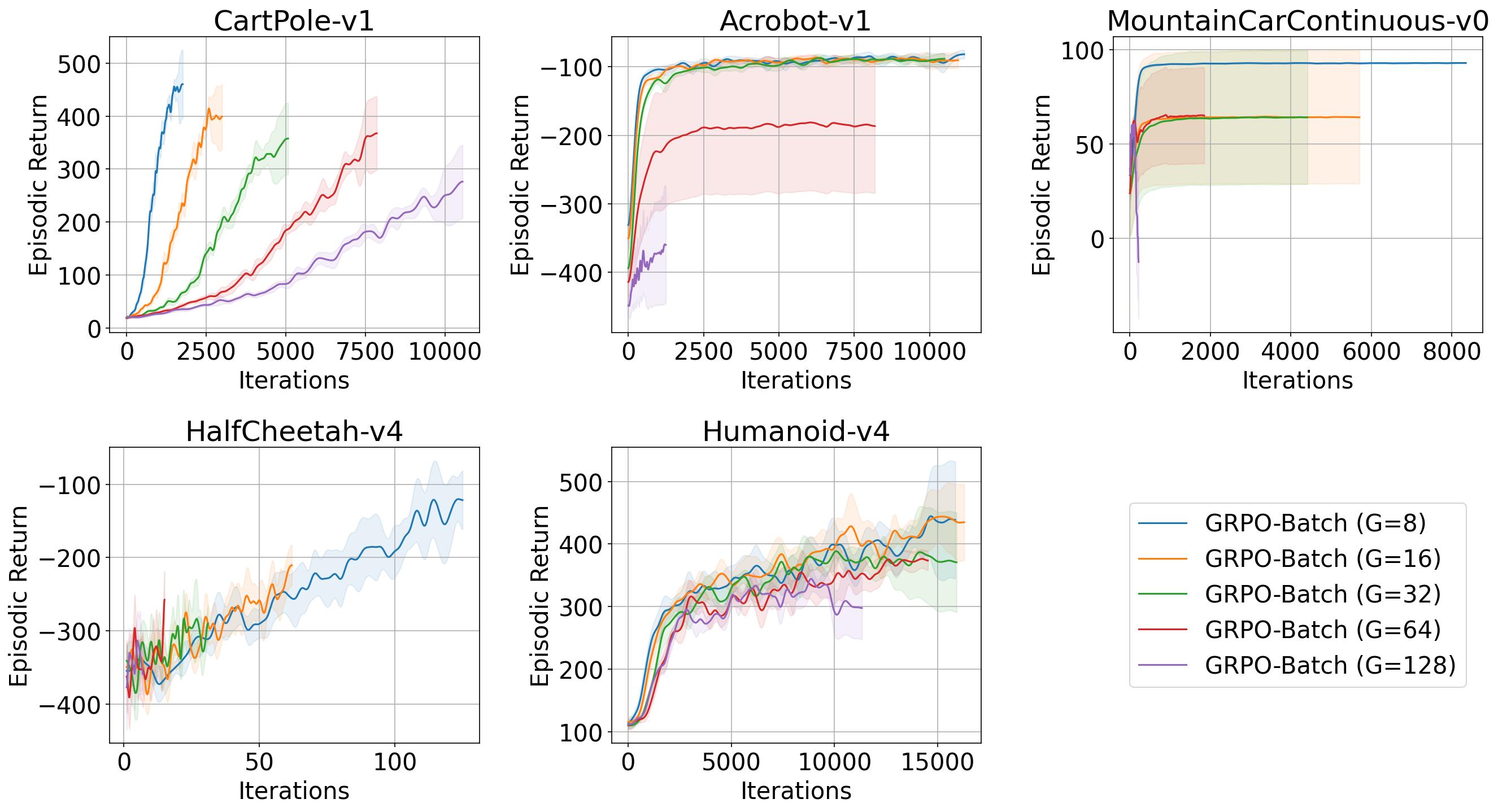}
    \caption{\textbf{Episodic return across training iterations with $\gamma=0.99$.}}
    \label{fig:group_size_h0_y0.99_iterations}
\end{figure}

\begin{figure}[h!]
    \centering
    \includegraphics[width=\linewidth]{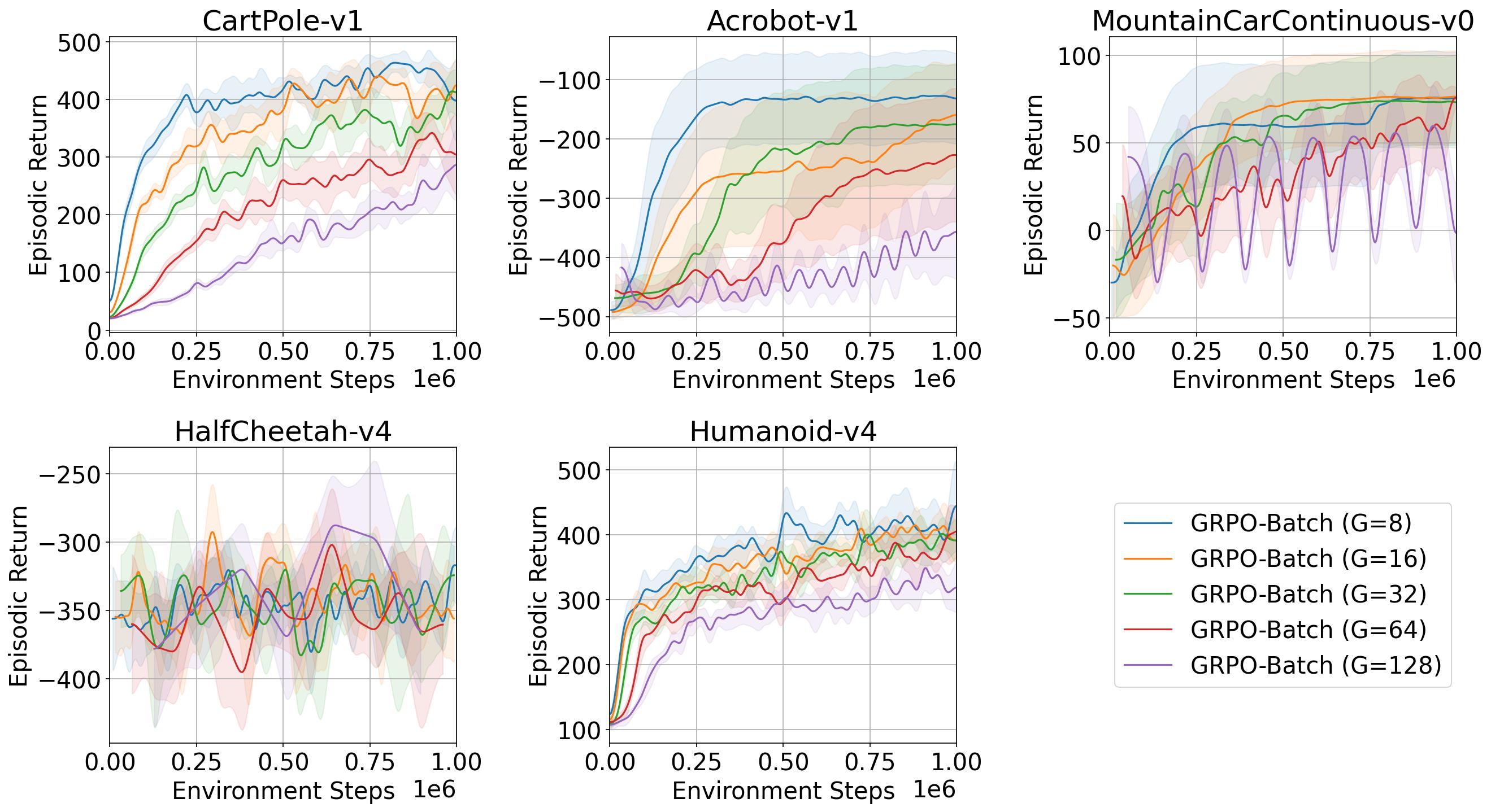}
    \caption{\textbf{Effect of group size across environments with $\gamma=1$.}}
    \label{fig:group_size_h0_y1}
\end{figure}

\begin{figure}[h!]
    \centering
    \includegraphics[width=\linewidth]{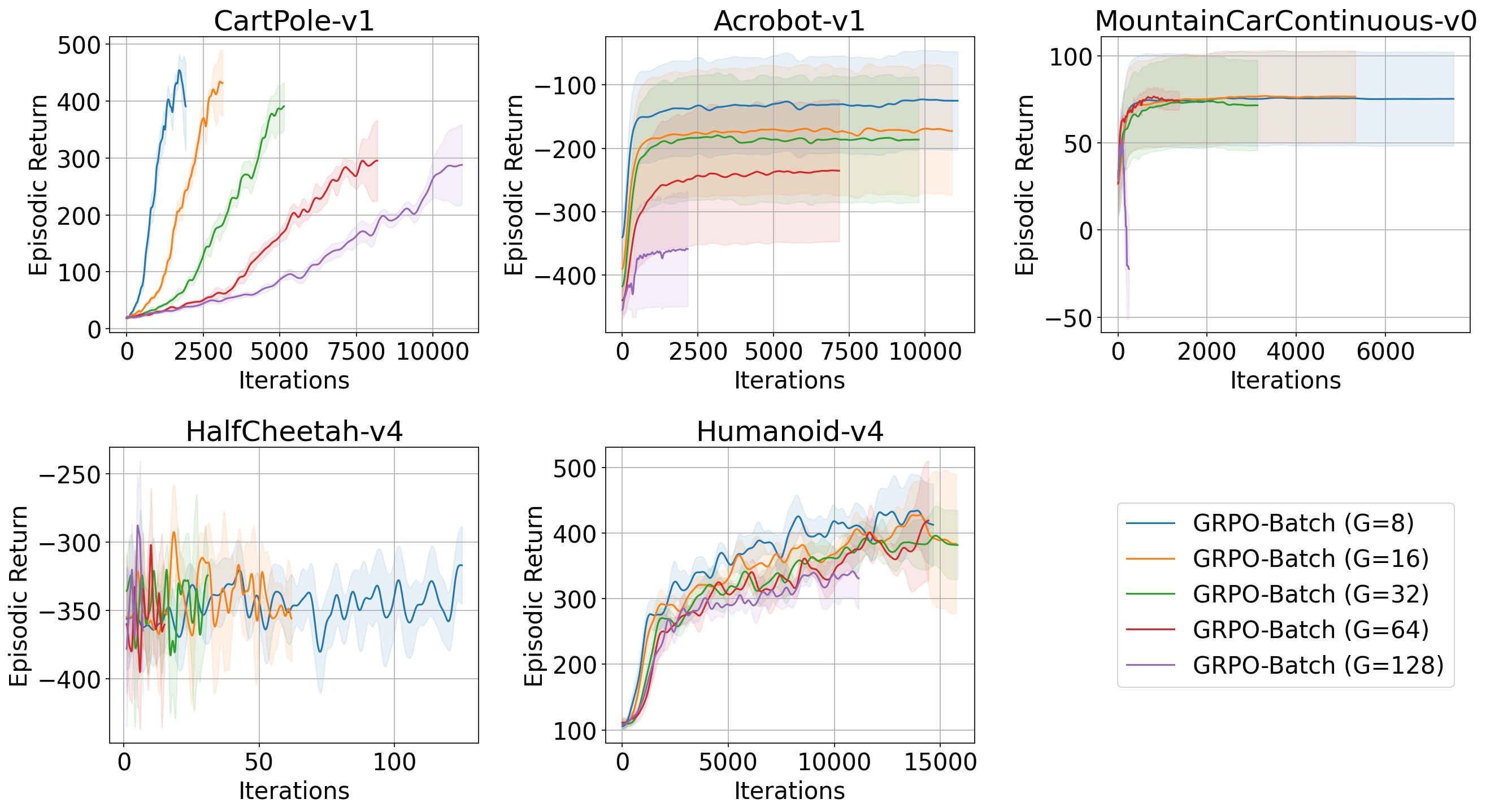}
    \caption{\textbf{Episodic return across training iterations with $\gamma=1$.}}
    \label{fig:group_size_h0_y1_iterations}
\end{figure}

%%%%%%%%%%%%%%%%%%%%%%%%%%%%%%%%%%%%%%%%%%%%%%%%%%%%%%%%%%%%

\end{document}